\documentclass[3p]{elsarticle} % or preprint
\pdfoutput=1

\usepackage{amsmath,amssymb,amsfonts}
\usepackage{algorithm}
\usepackage[noend]{algpseudocode}
\usepackage{graphicx}
\usepackage{textcomp}
\usepackage{bm}
\usepackage{subfig}
\usepackage{multirow}
\usepackage{threeparttable}
\usepackage{adjustbox}
\usepackage{stfloats}
\usepackage{tabularx,booktabs}
\usepackage{amsmath}
\usepackage{soul}
\usepackage{verbatimbox}
\newcolumntype{C}{>{\centering\arraybackslash}X} % centered version of "X" type
\def\BibTeX{{\rm B\kern-.05em{\sc i\kern-.025em b}\kern-.08em
    T\kern-.1667em\lower.7ex\hbox{E}\kern-.125emX}}
\usepackage{xcolor}
\usepackage{listings}
\newcommand\YAMLcolonstyle{\color{red}\mdseries}
\newcommand\YAMLkeystyle{\color{black}\bfseries}
\newcommand\YAMLvaluestyle{\color{blue}\mdseries}

\makeatletter
\def\verbatim@font{\linespread{1}\normalfont\ttfamily}

\newcommand\language@yaml{yaml}

\expandafter\expandafter\expandafter\lstdefinelanguage\expandafter{\language@yaml}
{
  keywords={true,false,null,y,n},
  keywordstyle=\color{darkgray}\bfseries,
  basicstyle=\YAMLkeystyle,                                 % assuming a key comes first
  sensitive=false,
  comment=[l]{\#},
  morecomment=[s]{/*}{*/},
  commentstyle=\color{purple}\ttfamily,
  stringstyle=\YAMLvaluestyle\ttfamily,
  moredelim=[l][\color{orange}]{\&},
  moredelim=[l][\color{magenta}]{*},
  moredelim=**[il][\YAMLcolonstyle{:}\YAMLvaluestyle]{:},   % switch to value style at :
  morestring=[b]',
  morestring=[b]",
  literate =    {---}{{\ProcessThreeDashes}}3
                {>}{{\textcolor{red}\textgreater}}1     
                {|}{{\textcolor{red}\textbar}}1 
                {\ -\ }{{\mdseries\ -\ }}3,
}

\lst@AddToHook{EveryLine}{\ifx\lst@language\language@yaml\YAMLkeystyle\fi}
\makeatother

\newcommand\ProcessThreeDashes{\llap{\color{cyan}\mdseries-{-}-}}

\makeatother

\journal{AMDL special issue of the MICPRO journal}

\begin{document}

\begin{frontmatter}

\title{Quantization of Deep Neural Networks for Accumulator-constrained Processors}
\tnotetext[mytitlenote]{This paper is an extended version of \cite{8491840}.}
\tnotetext[mytitlenote]{Email addresses: \texttt{e.d.bruin@tue.nl} (Barry de Bruin, corresponding author), \texttt{zoran.zivkovic@intel.com} (Zoran Zivkovic), \texttt{h.corporaal@tue.nl} (Henk Corporaal).}

\author[add1]{Barry de Bruin}
%\ead{e.d.bruin@tue.nl}
\author[add2]{Zoran Zivkovic}
%\ead{zoran.zivkovic@intel.com}
\author[add1]{Henk Corporaal}
%\ead{h.corporaal@tue.nl}

%% Include affiliations
\address[add1]{Eindhoven University of Technology; Eindhoven, The Netherlands}
\address[add2]{Intel Benelux; Eindhoven, The Netherlands}

\begin{abstract}
\iffalse
Artificial Neural Networks (NNs) can effectively be used to solve many classification and regression problems, and deliver state-of-the-art performance in the application domains of natural language processing (NLP) and computer vision (CV). However, the tremendous amount of data movement and excessive convolutional workload of these networks hampers large-scale mobile and embedded productization. Therefore these models are generally mapped to energy-efficient accelerators without floating-point support. Weight and data quantization is an effective way to deploy high-precision models to efficient integer-based platforms. In this paper a quantization method for platforms without wide accumulation registers is being proposed. Two constraints to maximize the bit width of weights and input data for a given accumulator size are introduced. These constraints exploit knowledge about the weight and data distribution of individual layers. Using these constraints, we propose a layer-wise quantization heuristic to find a good fixed-point network approximation. To reduce the number of configurations to consider, only solutions that fully utilize the available accumulator bits are being tested. We demonstrate that 16-bit accumulators are able to obtain a Top-1 classification accuracy within 1\% of the floating-point baselines on the CIFAR-10 and ILSVRC2012 image classification benchmarks.
\fi

We introduce an Artificial Neural Network (ANN) quantization methodology for platforms without wide accumulation registers. This enables fixed-point model deployment on embedded compute platforms that are not specifically designed for large kernel computations (i.e. accumulator-constrained processors). We formulate the quantization problem as a function of accumulator size, and aim to maximize the model accuracy by maximizing bit width of input data and weights. To reduce the number of configurations to consider, only solutions that fully utilize the available accumulator bits are being tested. We demonstrate that 16-bit accumulators are able to obtain a classification accuracy within 1\% of the floating-point baselines on the CIFAR-10 and ILSVRC2012 image classification benchmarks. Additionally, a near-optimal $2\times$ speedup is obtained on an ARM processor, by exploiting 16-bit accumulators for image classification on the All-CNN-C and AlexNet networks.

\end{abstract}

\begin{keyword}
quantization, fixed-point, efficient inference, narrow accumulators, convolutional neural networks
\end{keyword}

\end{frontmatter}

% \linenumbers

% start here
\section{Introduction}
Neural Networks (NNs) are a class of machine-learning algorithms that deliver state-of-the-art performance on many natural language processing (NLP) (e.g. speech recognition and natural language understanding) and Computer Vision (CV) tasks, such as object localization, classification, and recognition of objects. Unfortunately, both the training and inference phase of the NNs are computationally demanding. As a result, NNs are usually developed and trained on high-performance clusters, whereafter the resulting model (learned weights) is mapped to an optimized hardware platform for efficient model deployment on mobile and embedded devices. These hardware platforms generally use reduced-precision integer arithmetic, which simplifies the data path and reduces memory storage, bandwidth, and energy requirements.

The process of converting a pre-trained floating-point NN model to reduced-precision is called quantization. Typical quantization procedures for NNs check a number of possible reduced-precision solutions for both weights and intermediate data within a given network, and choose the cheapest minimal bit width solution within a tolerable model accuracy penalty\cite{DBLP:journals/corr/GyselMG16,Guo2017}. Due to the enormous set of potential quantization solutions, it is not feasible to find an optimal solution. Typically only a small number of solutions are evaluated. 

We identify that the bit width of partial result accumulators is generally a bottleneck on platforms that are not specifically optimized for large kernel computations, consisting of a large sequence of convolution operations. Kernel computations are the core of many NN architectures, such as traditional fully-connected NNs, Convolutional Neural Networks (CNNs) and Recurrent Neural Networks (RNNs). 

This accumulator bottleneck further complicates the quantization procedure. To address this issue, we formulate the NN quantization problem as a function of accumulator size. For a fixed accumulator bit width, we aim to maximize the model accuracy by maximizing the bit width of input data and weights. To increase the maximum data and weight bit width, we introduce two constraints that provide a more optimistic  maximum accumulator range estimate, while still providing analytical guarantees on avoiding potential accumulator overflow. 

This paper extends upon a previous work\cite{8491840} by including model finetuning and an evaluation on a real platform. The main contributions of this paper are:
%\iffalse
\begin{itemize}
\item A new quantization approach that considers a limited accumulator size. This is very useful for platforms with narrow accumulators. It includes:
\begin{itemize}
\item A quantization method to maximize the bit width of weights and input data for a given accumulator size within an NN layer.
\item A heuristic for fast layer-wise quantization of complete CNNs for image classification.
\item A novel fixed-point finetuning method that improves the quantization solution for accumulator-constrained platforms.
\end{itemize}
\item An evaluation of our quantization approach on three popular CNN benchmarks, including performance benchmarks on a representative platform.
%\item A performance evaluation of the obtained solutions on a representative platform.
\end{itemize}
%\fi

%The main contributions of this paper are as follows: Firstly, we formulate the NN quantization problem as a function of accumulator size. For a fixed accumulator bit width, we aim to maximize the model accuracy by maximizing the bit width of input data and weights. To increase the maximum data and weight bit width, we introduce two constraints that reduce the number of additional accumulation bits. These constraints do still provide analytical guarantees on avoiding potential accumulator overflow. Secondly, we provide a heuristic for fast quantization of complete CNNs for image classification by only considering solutions that fully utilize all available accumulator bits. Finally, we evaluate our layer-wise quantization heuristic on three popular CNN benchmarks and demonstrate that competitive classification accuracy is attainable with narrow accumulators of only 16-bit.

The rest of the paper is organized as follows: Section 2 summarizes related work in CNN quantization. Section 3 covers preliminaries and introduces notation and background for fixed-point inference. Section 4 introduces the quantization method for accumulator-constrained accelerators. Section 5 explains the heuristic for layer-wise CNN quantization. Section 6 explains the accumulator-constrained finetuning approach. Evaluation and experimental results are provided in Section 7. Concluding remarks follow in Section 8.

% two comments still left:
% - better term for 'constraint'
% - better definition of 'the impact of large kernel computations on the required data path bit width'

\section{Related Works}
Recent works have investigated the feasibility of converting floating-point NNs to fixed-point for efficient implementation onto mobile and embedded devices. It has been well-established that NNs are generally very resilient to quantization noise and do not require large bit widths for good performance. Courbariaux, David and Bengio\cite{Courbariaux2014} train a CIFAR-10 network with a negligible accuracy penalty, using only 10 bits for weights and activations (and 12 bits for weight updates). A primary innovation is the application of a dynamically scaled fixed-point format. Their training procedure monitors if overflow happens, and adjusts the scaling factors of weights and data in NN layers accordingly. FlexPoint\cite{flexpoint} extends this work by providing hardware support for dynamic overflow-based scaling factor management. Their custom 16-bit floating-point format with shared exponents matches the classification accuracy of 32-bit floating-point baselines on several large NN benchmarks. Training with 16-bit integers and a 32-bit accumulator was further explored by Das et al.\cite{DBLP:journals/corr/abs-1802-00930}. To prevent accumulator overflow, the kernel computation is blocked and partial results are added to a floating-point accumulator. To reduce the integer to float conversion overhead, precision of input data is reduced to allow for larger partial result accumulation chains. A very recent work evaluates the impact of reduced-precision floating-point accumulators for reduced-precision training\cite{DBLP:journals/corr/abs-1901-06588}. Rastegari et al.\cite{journals/corr/RastegariORF16} obtains respectable classification accuracy on the difficult ILSVRC2012 benchmark with only 1-bit weights.

%Gupta et al.\cite{Gupta:2015:DLL:3045118.3045303} proposes a method to train CNNs using 16-bit fixed-point integers. The main contribution of this work is the use of a stochastic rounding scheme to prevent the update steps from getting lost. 

The downside of above-mentioned quantization approaches is that the training procedure of large NNs can be very time-consuming. For many use cases it suffices to retrain a pre-trained model on your own (similar) dataset\cite{10.1007/978-3-319-10590-1_53}. Many related works\cite{DBLP:journals/corr/GyselMG16, Lin2016, Guo2017} do therefore focus on quantization of a pre-trained model within a tolerable accuracy penalty. 
%To regain some of the lost accuracy during quantization, the quantized model is generally finetuned (retrained)\cite{DBLP:journals/corr/GyselMG16, Anwar2015, Lin2016, Guo2017}.

Vanhoucke et al.\cite{37631} linearly normalizes weights and (sigmoid) activations of every layer in a speed-recognition NN to 8-bit by analysing the range of weights and activations. A similar approach is implemented in several deep learning frameworks such as Tensorflow\cite{tensorflow2015-whitepaper} and Caffe-Ristretto\cite{Gysel2018}. Lin, Talathi, and Annapureddy\cite{Lin2016} propose an analytical model to quickly convert pre-trained models to fixed-point. The advantage of this model is that it does not require exhaustive layer-wise optimization, but rather determines the bit width of every layer based on the approximate parameter and data range distribution.

An exhaustive layer-wise optimization is a straight-forward approach for NN quantization. This procedure generally consists of testing many possible quantization solutions for every layer in the network\cite{Guo2017, Anwar2015}. To reduce the number of solutions to consider, several heuristics were developed. Gysel et al.\cite{Gysel2018} propose an iterative quantization procedure where weights are quantized first, and activations are quantized second. A similar two-step approach is described by other related works\cite{Qiu:2016:GDE:2847263.2847265}. Shan et al.\cite{Shan2016} considers some target platform characteristics, and shows that the accumulator bit width can be reduced to 16-bit without a significant penalty in classification accuracy for the LeNet5 benchmark.

To regain some of the lost accuracy during quantization, the quantized model is generally finetuned (retrained)\cite{DBLP:journals/corr/GyselMG16, Anwar2015, Lin2016, Guo2017}. The primary difficulty in fixed-point retraining lies in the update step of the Stochastic Gradient Descent (SGD) algorithm, which iteratively updates all weights by a fraction of its output loss contribution. With insufficient weight precision, the update steps get quantized to zero, which prohibits learning. A common training and finetuning procedure\cite{Courbariaux2014} is to perform the forward and gradient computation in reduced precision, while the weights are updated in high-precision\cite{Anwar2015, DBLP:journals/corr/GyselMG16}. After training these high-precision weights are permanently quantized, which allows for efficient model deployment.

\iffalse
More aggressive fixed-point training, up to the point where both the weights and activations are binarized during forward and backward computation\cite{journals/corr/RastegariORF16} with respectable performance on difficult benchmarks\cite{lin2017towards}. These methods still require high-precision weights during training. However, at test-time the whole network can be binarized.
\fi

We consider a target platform that is not specifically optimized for large NN kernels and does not have a wide accumulation data path. Knowing that the accumulator is the bottleneck, we define several constraints that greatly reduce the number of solutions to consider. The quantization problem is reformulated to maximize the model accuracy for a fixed accumulator bit width. We adopt fixed-point finetuning using high-precision weights. Dynamic fixed-point scaling is applied to prevent the accumulator from overflowing during finetuning. However, during test-time the whole network will be quantized to the derived bit widths. 

Differently from\cite{DBLP:journals/corr/abs-1802-00930}, we do not split the kernel computation, thereby facilitating optimal throughput of kernel computations. Additionally, our methodology is applicable to pre-trained models for difficult image classification benchmarks, which adds more practical value.

%\newpage

\section{Background}
This section covers NN preliminaries and introduces notation and relevant background for fixed-point NN inference.

\subsection{Convolutional Neural Networks}
Convolutional Neural Networks (CNNs) are a class of NNs that work very well for a variety of Computer Vision tasks. They can be used as an end-to-end approach for object classification, localization, tracking, or even higher-order tasks such as object recognition. CNNs are composed of layers that are generally connected in a feed-forward fashion. The input of the first layer is an (preprocessed) image. The model output depends on the task, and can be a class likelihood vector for image classification, a bounding box for object localization, or a combination of both.
\begin{figure}[htbp]
%%\centerline{\includegraphics[width=90mm,keepaspectratio]{visualization/figures_eps/cnn_layers_14.eps}}
\centerline{\includegraphics[width=90mm,keepaspectratio]{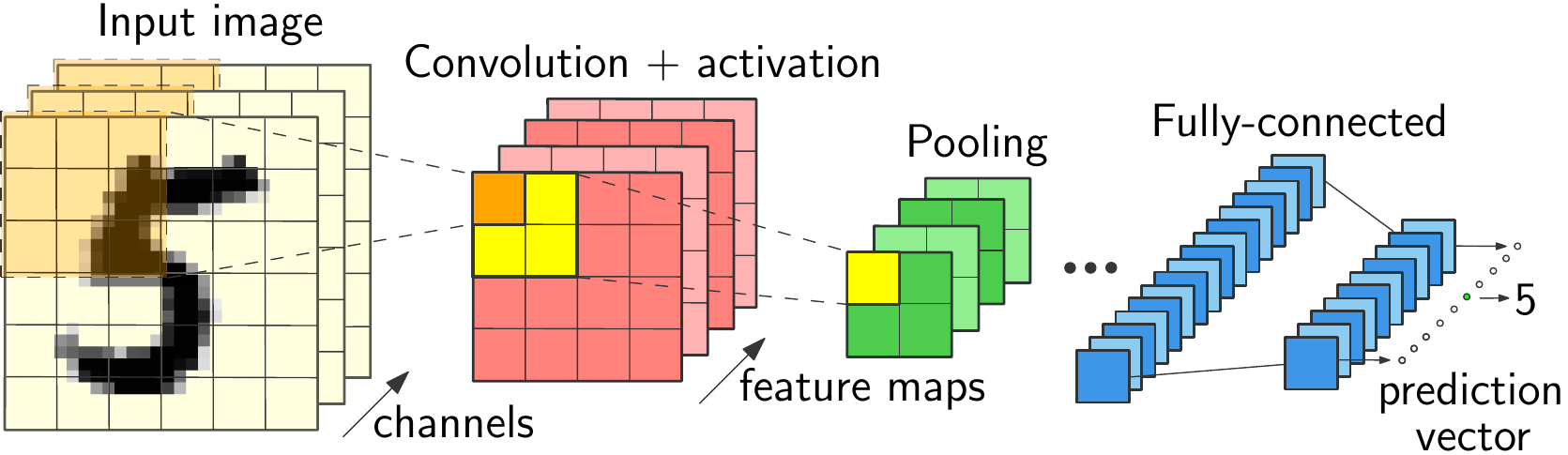}}
\caption{Example CNN for handwritten digit classification.}
\label{Figure:CNN}
\end{figure}

A typical CNN consists of several stacks of convolution + activation + pooling (+ normalization) layers for feature extraction, followed up by several fully-connected (+ activation) layers for classification. The main computations of a CNN are the convolution and fully-connected layers. Both layer types can be simplified to a series of multiply-accumulate operations or dot product:
\begin{equation}
	y = \sum^K_{i=1} w_i d_i
\end{equation}
where $w_i$ are the learned weights (and bias), $d_i$ are data values, $y$ is the output result of a single neuron before activation, and $K$ is the kernel size.

\subsection{Fixed-point preliminaries}
NN models are generally trained using the single-precision floating-point format. To approximate these models on a target platform that only supports 2's complement integer arithmetic, we use a fixed-point number representation. This format can be represented as a combination of Bit Width ($BW$), Integer (Word) Length ($IL$), Fractional (Word) Length ($FL$), and a sign bit:
\begin{equation}
\label{eqn:4}
	BW_{\bm{x}} = IL_{\bm{x}} + FL_{\bm{x}} + 1
\end{equation}
where $\bm{x}$ corresponds to a set of floating-point values that share the same fixed-point format. An example is depicted in Figure \ref{Figure:num_ex} below: 
\begin{figure}[htbp]
\centerline{\includegraphics[width=90mm,keepaspectratio]{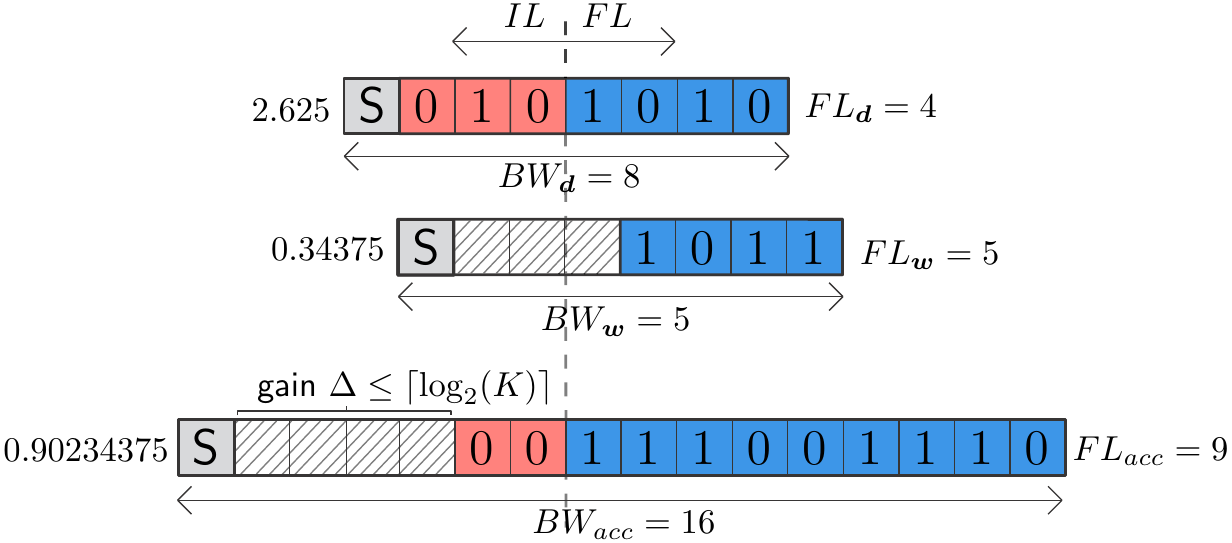}}
%\centerline{\includegraphics[width=90mm,keepaspectratio]{visualization/figures_eps/data_path_3.eps}}
\caption{Numerical example of fixed-point multiply-accumulation.}
\label{Figure:num_ex}
\end{figure}

\noindent
Any real value $x \in \bm{x}$ can be quantized to its integer representation by scaling and rounding:
\begin{equation}
    Q(x) = round(x\cdot 2^{FL_{\bm{x}}})
\end{equation}
The original real value can be estimated by scaling the integer values back:
\begin{equation}
\label{eqn:quantized}
    \hat{x} = Q(x)\cdot 2^{-FL_{\bm{x}}}
\end{equation}
The maximum quantization error is bounded by the choice of rounding function. For example, if we consider a round-to-nearest policy, the quantization error is bounded by half of the least significant digit e.g. $2^{-(FL_{\bm{x}} +1)}$. To minimize this quantization error the fractional length or scaling factor should be maximized. However, increasing the fractional length of a fixed-point format will reduce the representable range:
\begin{equation}
\label{eqn:1}
     -2^{IL_{\bm{x}}}\leq \hat{x} \leq 2^{IL_{\bm{x}}} - 2^{-FL_{\bm{x}}}
\end{equation}
If data values fall outside the range of our 2's complement integer representation, we either need to clip these values or overflow will happen. Therefore we must ensure that the integer length is chosen large enough to prevent overflow:
\begin{equation}
\label{eqn:3}
	IL_{\bm{x}} = \lfloor \log_2 \left( R_{\bm{x}} \right) \rfloor + 1
\end{equation}
where range $R_{\bm{x}}$ corresponds to the absolute maximum value within set ${\bm{x}}$ i.e.
\begin{equation}
\label{eqn:2}
 	R_{\bm{x}} = \max_{x \in \bm{x}} |x|
\end{equation}
For example, consider a fixed-point group ${\bm{x}}$ with $BW_{\bm{x}}=16$ and a maximum range of $R_{\bm{x}}=0.1256$. From Equation \ref{eqn:3} follows that we require at least $IL_{\bm{x}}=-2$ to prevent overflow. Stated differently, maximum precision can be obtained by scaling all values in group ${\bm{x}}$ by $2^{FL_{\bm{x}}}$ for $FL_{\bm{x}} = 17$, using Equation \ref{eqn:4}.

The range estimate of Equation \ref{eqn:2} has been used by others\cite{DBLP:journals/corr/GyselMG16}, and worked very well for our experiments. Other\cite{Kim1994, Lin2016}, more optimistic, range estimates were also considered, but without much success; results of different range estimates were inconsistent between different benchmarks. These findings indicate that having sufficient range is more important than precision, which is in line with the work of Lai et al.\cite{DBLP:journals/corr/LaiSC17}.

\subsection{Range Analysis}
\begin{figure}[htbp]
\centerline{\includegraphics[width=90mm,keepaspectratio]{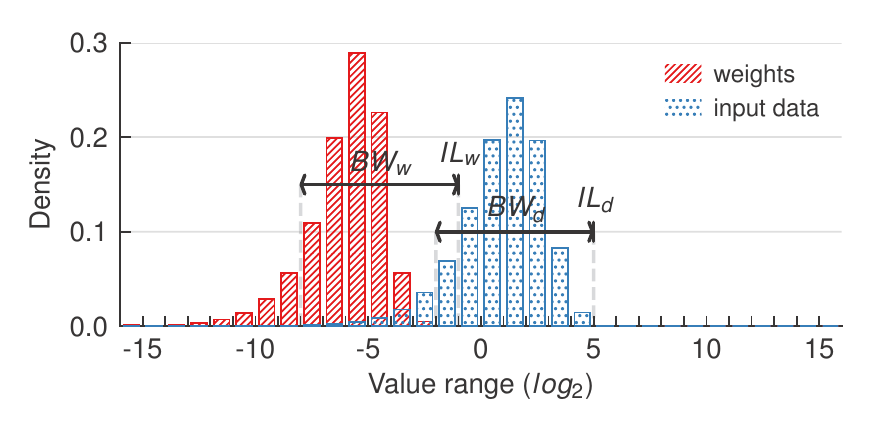}}
%\centerline{\includegraphics[width=90mm,keepaspectratio]{{visualization/range_features6-eps-converted-to.pdf}}}
\vspace{-0.5cm}
\caption{Value range of weights (red) and input data (blue) of AlexNet's third convolution layer. The two intervals suggest a potential quantization solution. Values outside the intervals cannot be represented. Zero-valued data is ignored in this distribution.}
\label{Figure:alexnet_conv3}
%  Zero can always be represented (see Equation \ref{eqn:1}).
\end{figure}

\begin{figure}[htbp]
\centerline{\includegraphics[width=90mm,keepaspectratio]{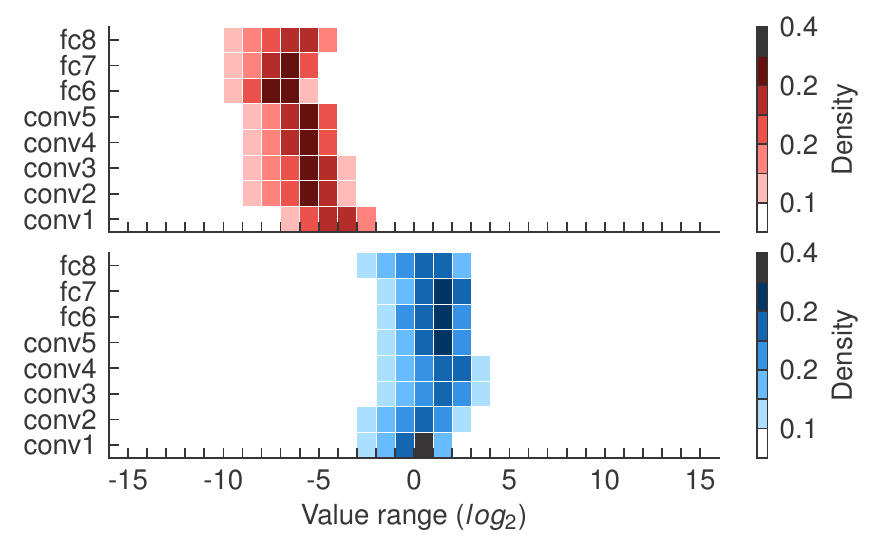}}
%%\vspace{-0.5cm}
\caption{Value range of weights (top) and input data (bottom) of AlexNet's convolutional and fully-connected layers (conv3 corresponds to Figure \ref{Figure:alexnet_conv3}).}
\label{Figure:alexnet_heatmap}
\end{figure}

A common approach to determine $R_{\bm{x}}$ is to analyse the range of weights and input data\cite{Shan2016, Guo2017, Gysel2018, tensorflow2015-whitepaper} for every convolutional or fully-connected layer within a NN. The weight range is extracted from the high-precision pre-trained model. The data range can be estimated by forwarding a large batch of images through the network. Figure \ref{Figure:alexnet_conv3} visualizes a typical density distribution of the weights and input values of a layer from the popular AlexNet\cite{Krizhevsky2014} network. It can be observed that weights are significantly smaller than input data, but that both groups are clustered together. Additionally, the range between layers also differs significantly, as is depicted in Figure \ref{Figure:alexnet_heatmap}. 

Using a single fixed-point format for the whole network is not optimal due to the large range difference between different groups and layers. To encode this network in fixed-point, we require an integer length that is large enough to prevent overflow, while the precision or fractional length should be large enough to distinguish variations in weights.

In NNs this dynamic range problem can be addressed by splitting weights and input data within a layer into separate \textit{fixed-point groups}\cite{Courbariaux2014}. A \textit{fixed-point group} is a set of values that share the same bit width and fractional length. In other words, the global scaling factor is replaced by multiple local scaling factors. An example of this is shown in Figure \ref{Figure:alexnet_conv3}, where both weights and data have a separate integer length and bit width to optimally capture the data range.

Using different fixed-point groups for both weights and data in layer adds minimal computational overhead. Weights can be quantized off-line and integer multiplication of two fixed-point groups with different scaling factors require no precision alignment. Additionally, we do not reduce precision of intermediate results during kernel computation, or utilize wider accumulators to store partial results\cite{DBLP:journals/corr/abs-1802-00930}, as is depicted in Figure \ref{fig:SP64} below.
\begin{figure}[h!]
%\begin{minipage}{1\textwidth}
%\centering % left, lower, right, upper
%\includegraphics[page=2,width=0.8\textwidth]{visualization/figures/data_path}
%\end{minipage}
\begin{minipage}{30mm}
\centering
\begin{tabular}{c}
\includegraphics[trim={0cm 0cm -0.6cm 0cm},width=30mm,keepaspectratio]{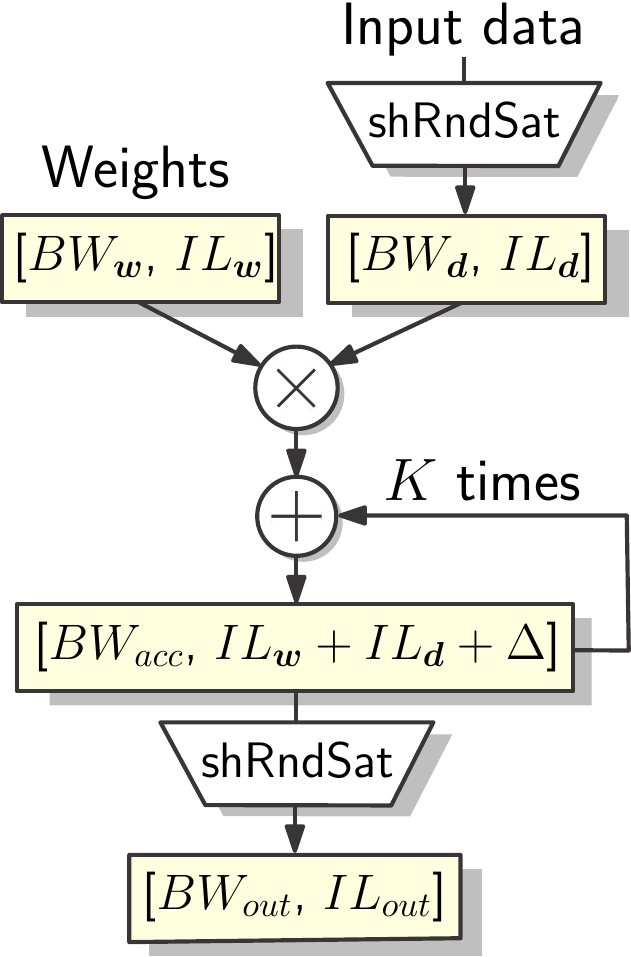}
\end{tabular}
\end{minipage}
\begin{minipage}{60mm}
\centering
%\begin{verbatim}
%\begin{verbbox}
\begin{adjustbox}{width=60mm}
\begin{lstlisting}[basicstyle=\linespread{1.1}\ttfamily,columns=fullflexible,keepspaces=true]
for (k,0,512) {
  acc[ramp(0,1,8)] += d[ramp(k*8,1,8)]* x8(w[k])
}

.LBB211_89:  # "for k"
  add      r3, r8, r0
  vld1.16  {d20, d21}, [r2:128]!   # q10 := d
  vld1.16  {d18[], d19[]}, [r3:16] # q9 := w
  add      r0, r0, #2
  cmp      r0, #1024
  vmla.i16 q8, q9, q10
  bne .LBB211_89
\end{lstlisting}
\end{adjustbox}
%\end{verbbox} %\end{verbatim}
%\resizebox{62.5mm}{!}{\theverbbox}
\end{minipage}
\caption{Illustration of vectorized fixed-point convolution with 16-bit accumulators including some reference code for the evaluated ARM platform (Section \ref{sec:arm_proc}). \textit{shRndSat} introduces some rescaling overhead between layers. Also, there is no rescaling in the kernel computation required.}
\label{fig:SP64}
\end{figure}

Results of a previous layer that are used as input for the next layer might require precision alignment. This \textit{shRndSat} procedure consists of a single arithmetic shift with clipping and rounding (21 instructions for 8$\times$16-bit integers on the evaluated ARM platform), which is negligible, when we consider that every input sample can generally be reused for 100--1000s multiply-accumulations.

\section{Quantization with narrow accumulators}
\label{sec:qwna}
In the previous section we introduced the relevant concepts for fixed-point NN inference and motivated the use of multiple fixed-point groups. In this section we define our quantization methodology for convolutional and fully-connected layers for a data path with limited data bus and accumulator size.
\iffalse
The goal of the quantization procedure is to find a fixed-point format for both parameters and input data. A generic quantization approach of a convolutional or fully-connected layer can be summarized by the following steps:
\begin{enumerate}
\item Analyse the parameter and input data range to determine $IL_{\bm{p}}$ and $IL_{\bm{d}}$. The available bit widths $BW_{\bm{p}}$ and $BW_{\bm{d}}$ determine the precision $FL_{\bm{w}}$ and $FL_{\bm{d}}$.
\item Check if the chosen solution does not overflow the accumulator. The accumulated result $y$ will require a worst-case bit width $BW_{\bm{p}} + BW_{\bm{d}} + \lceil \log_2 K \rceil$.
\item Adapt the layer to the new configuration, and check if the network performance is still sufficient.
\end{enumerate}
To test which combination of $BW_{\bm{w}}$ and $BW_{\bm{d}}$ provides the
best trade-off between performance and model size, only a few solutions have to be tested, given that the accumulator is large enough. For example, if we consider a platform with support for 8 and 16-bit integer data types, only 4 combinations need to be tested. Other options would be suboptimal, as these either do not fully utilize all bits, or require a data type that is not natively supported by the target platform. However, if the accumulator is a bottleneck, we must decrease the bit width of groups ${\bm{w}}$ and ${\bm{d}}$ until the accumulator will not overflow. Testing every combination of $BW_{\bm{w}}$ and $BW_{\bm{d}}$ would be very time-consuming. 
\fi

%\iffalse
\label{sec:constraints}
Consider a target platform with a data bus bit width $BW_{data}$ and accumulator bit width $BW_{acc}$, as visualized in Figure \ref{fig:SP64}. These platform-dependent parameters restrict the bit width of weights ($BW_{\bm{w}}$) and input data ($BW_{\bm{d}}$):
\begin{equation}
	1\leq BW_{\bm{w}}\leq BW_{data}\qquad 1\leq BW_{\bm{d}}\leq BW_{data}
\end{equation}
For efficiency reasons we do not consider sizes that exceed the data bus width. Combinations of $BW_{\bm{w}}$ and $BW_{\bm{d}}$ are also bounded by the accumulator size:
\begin{equation}
	\label{eqn:10}
	(BW_{\bm{w}} + BW_{\bm{d}} - 1) + \Delta \leq BW_{acc}
\end{equation}
where $\Delta$ is a layer-dependent parameter that corresponds to the number of additional integer bits the accumulator requires to store the kernel computation.

A generic quantization approach of a convolutional or fully-connected layer can be summarized by the following steps\cite{DBLP:journals/corr/GyselMG16, Lin2016, Guo2017}:
\begin{enumerate}
\item Analyse the parameter and input data range to determine integer lengths $IL_{\bm{w}}$ and $IL_{\bm{d}}$. The available bit widths $BW_{\bm{w}}$ and $BW_{\bm{d}}$ determine the precision $FL_{\bm{w}}$ and $FL_{\bm{d}}$ (see Equation \ref{eqn:4}).
\item Check if the chosen solution does not overflow the accumulator (see Equation \ref{eqn:10}).
\item Adapt the layer to the new configuration, and check if the network performance is still sufficient.
\end{enumerate}
To test which combination of $BW_{\bm{w}}$ and $BW_{\bm{d}}$ provides the
best model accuracy, we must decrease the bit width of groups ${\bm{w}}$ and ${\bm{d}}$ until the accumulator will not overflow. Testing every combination of $BW_{\bm{w}}$ and $BW_{\bm{d}}$ would be very time-consuming. Instead, we define several constraints that provide an upper bound on the maximum accumulator range.

\begin{figure}[htbp]
\centerline{\includegraphics[width=90mm,keepaspectratio]{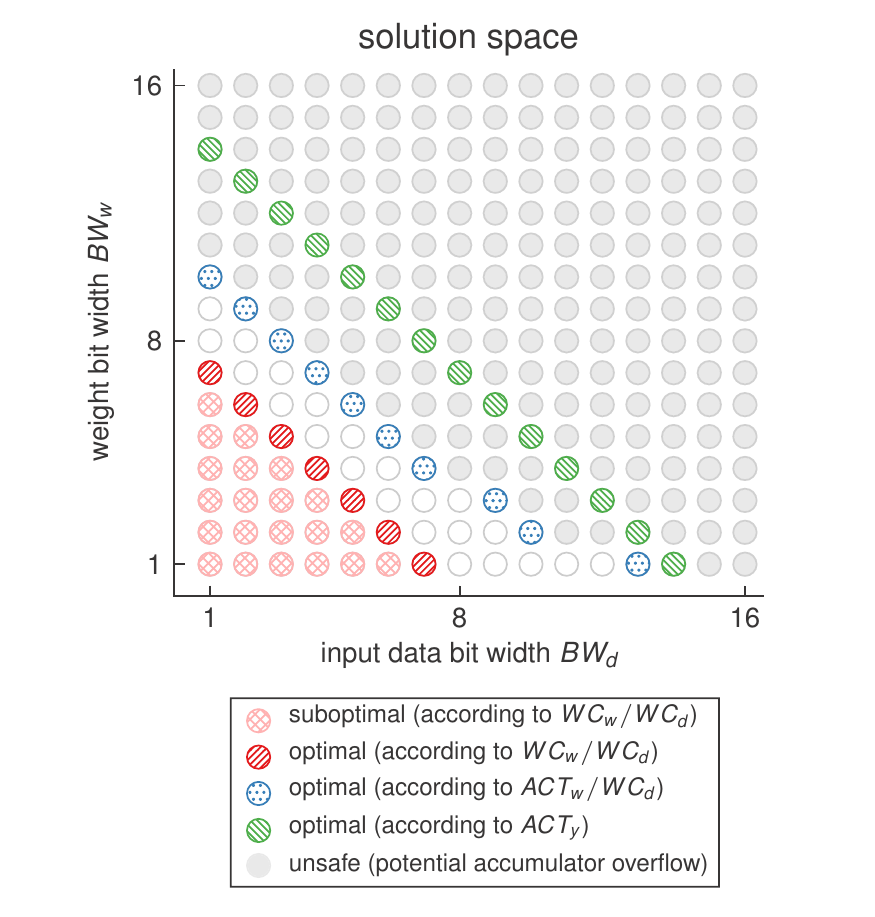}}
%%\centerline{\includegraphics[width=90mm,keepaspectratio]{visualization/solution_space_example-eps-converted-to.pdf}}
\caption{Illustration of the solution space of a convolutional or fully-connected layer. Red dots are solutions that utilize all accumulator bits for the worst-case constraint. Light red dots are suboptimal as these do not utilize all accumulator bits. Blue dots represent optimal solutions for the $ACT_w/WC_d$ constraint. For small bit widths most weights are quantized to zero, which explains the distortion at $BW_w < 4$.  Green represents the optimal solutions according to constraint $ACT_y$. However, these solutions may result in accumulator overflow.}
\label{Figure:solution_space}
\end{figure}

% BOUND 1
\subsection{Pessimistic constraint ($WC_w/WC_d$)} The most pessimistic constraint considers worst-case input and weight values for every multiply-accumulation. The accumulation register must be large enough to store $K$ kernel accumulations of $w_i d_i$ i.e.
\begin{equation}
\label{eqn:bbb}
	(BW_{\bm{w}} + BW_{\bm{d}} - 1) + \lceil \log_2 K \rceil \leq BW_{acc}
\end{equation}
Using Equation \ref{eqn:bbb} we can calculate which bit width combinations of weights and input data optimally utilize the accumulator:
\begin{equation}
\label{eqn:b1ddd}
	BW_{\bm{w}} + BW_{\bm{d}} = BW_{acc} + 1 - \lceil \log_2 K \rceil
\end{equation}
By only considering solutions that fully utilize the accumulator bit width, the number of quantization solutions that need to be tested per layer is reduced from a 2-dimensional search to an 1D search, and is usually very small. For example, if we consider a target platform with a 16-bit accumulator and a kernel $K = 5\times 5\times 16$, only a few configurations for $BW_{\bm{w}}$ and $BW_{\bm{d}}$ need to be considered, as is illustrated in Figure \ref{Figure:solution_space}.

% BOUND 2
\subsection{Conservative constraint ($ACT_w/WC_d$)} By using knowledge about our kernel weights, we can derive a more optimistic bound on the maximum accumulator range $R_{acc}$\cite{Yates2010}, using the fact that the sum of products is at most equal to the sum of absolute products:
\begin{equation}
	R_{acc} = \sum_{i=1}^K w_i d_i \leq  \sum_{i=1}^K |w_i| |d_i|
\end{equation}

To guarantee that this bound will not overflow the accumulator, we need to compute the maximum kernel range including quantization effects (i.e. rounding errors). Additionally, we need to assume worst-case input data for every multiply-accumulate.
\iffalse
\begin{equation}
	R_{acc} = \sum_{i=1}^K |\hat{w_i}|\cdot 2^{IL_{\bm{d}}}
\end{equation}
\fi
This results in a simple definition of the maximum accumulator range:
\begin{equation}
	R_{acc} = R_{kernel}\cdot 2^{IL_{\bm{d}}}
\end{equation}
where kernel range $R_{kernel}$ corresponds to the maximum sum of absolute quantized weights over all kernels within an NN layer:
\label{eqn:range_kernel}
\begin{equation}
	R_{kernel} = \max\limits_{kernels\ \in\ layer}\left(\sum_{i=1}^K |\hat{w_i}|\right)
\end{equation}
$R_{kernel}$ depends on the scaling factor $FL_{\bm{w}}$ and rounding policy (see Equation \ref{eqn:quantized}). We assume a round-away-from-zero tie-breaking policy, and pre-compute a table with the kernel range $R_{kernel}$ as a function of $FL_{\bm{w}}$ for a given layer.

We will now use $R_{acc}$ to compute $IL_{acc}$ (Equation \ref{eqn:3}). Using $IL_{acc}$ we can now derive our accumulator constraint $BW_{acc}$ where we assume that the precision of products is not being reduced during kernel computation:
\begin{align}
	BW_{acc} &= FL_{\bm{w}} + FL_{\bm{d}} + IL_{acc} + 1 \\
	       &= FL_{\bm{w}} + BW_{\bm{d}} + \lfloor \log_2 \left( R_{kernel} \right) \rfloor + 1 \nonumber \\ 
\intertext{Now replace $FL_{\bm{w}}$ by $BW_{\bm{w}}-IL_{\bm{w}}-1$ (Equation \ref{eqn:4}) and reorder the terms. This results in the following constraint:}
BW_{\bm{w}} + BW_{\bm{d}} &= BW_{acc} - \lfloor \log_2 \left( R_{kernel} \right) \rfloor + IL_{\bm{w}} 
\end{align}
This constraint will typically give more available bits than the $WC_w/WC_d$ constraint, as weights are generally very small. Since we did include worst-case representable input data and quantized kernel values, the accumulator will never overflow.

% BOUND 3
\subsection{Optimistic constraint ($ACT_{y}$)} An even more optimistic, but potentially unsafe, accumulator constraint can be derived by exploitation of the wrap-around property of 2's complement integer arithmetic. Note that this constraint does not prevent the intermediate accumulator result from overflowing. This is not an issue as long as the final output result fits within valid accumulator range. The maximum output range $IL_{\bm{y}}$ was already estimated during the range analysis step, which limits the precision of $FL_{\bm{w}}$ and $FL_{\bm{d}}$ as follows:
\begin{equation}
	BW_{acc} = FL_{\bm{w}} + FL_{\bm{d}} + IL_{\bm{y}} + 1
\end{equation}
With some rewriting this results in the following constraint for valid combinations of $BW_{\bm{w}}$ and $BW_{\bm{d}}$:
\begin{equation}
	BW_{\bm{w}} + BW_{\bm{d}} = BW_{acc} + 1 - \max(0, IL_{\bm{y}} - (IL_{\bm{w}} + IL_{\bm{d}}))
\end{equation}
It should be emphasized that the choice of $IL_{\bm{y}}$ is data-dependent and should therefore be chosen rather pessimistic to prevent accumulator overflow. The $\max$-operator is necessary to ensure that the result of a single multiplication will fit into the accumulator.  

In the next section we will extend this layer-wise quantization method with a heuristic to find a quantization solution for a complete network.
%\fi

\section{Heuristic layer-wise optimization}
\label{sec:heuristic}
This section describes the quantization heuristic for complete CNNs.
The goal of this procedure is to maximize the Top-1 classification accuracy for a fixed accumulator size and maximum data bit width.

Every convolution and fully-connected layer has a set of feasible quantization solutions. In Section \ref{sec:qwna} several accumulator constraints were introduced to reduce the solutions space for a given layer. However, testing every possible combination for every layer would still be very time-consuming. Instead we use a straightforward heuristic which iteratively quantizes the network: We start at the first eligible layer at the input of a pre-trained high-precision network, test all solutions that were proposed by the accumulator constraint, set the layer to the best solution for $BW_{\bm{w}}$ and $BW_{\bm{d}}$, and repeat this process for the next eligible layer. This process continues until all layers are quantized.

\begin{figure}[h]
\centerline{\includegraphics[width=90mm]{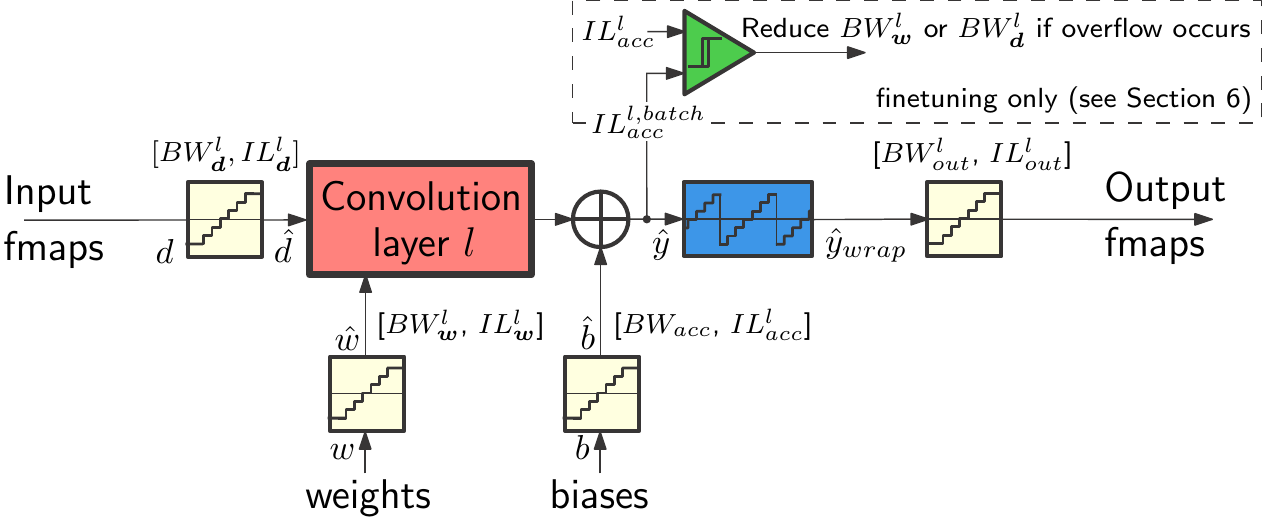}}
%%\vspace{-0.25cm}
\caption{Simulated fixed-point convolution or fully-connected layer during forward pass. Quantizers (yellow) reduce range and precision of high-precision input data, weights, biases and pre-activated output values. A wraparound operator simulates integer overflow (blue).}
\label{Figure:DP}
\end{figure}
Similar to other works\cite{Courbariaux2014, Gupta:2015:DLL:3045118.3045303}, we simulate a fixed-point data path with limited precision, as depicted in Figure \ref{Figure:DP}. This approach makes it very easy to set a subset of layers to reduced-precision, while the remainder stays high-precision. This implies that several quantization operators are inserted before and after every convolution and fully-connected layer. These uniform quantizers map the high-precision input data, weights and accumulator values to a discrete space. During the forward pass these quantizers will reduce the precision of fixed-point groups according to Equation \ref{eqn:quantized}. Range of input data and weights is clipped within representable range (i.e.\ Equation \ref{eqn:1}) using the integer lengths found during range analysis. To ensure correct operation, an integer accumulator is simulated by including a wrap-around operator after layer output $\hat{y}$:
\begin{equation}
	\hat{y}_{wrap} = y_{min} + (y_{min} + \hat{y})\ \mathbf{mod}\ y_{range}
	\label{Eq:wrap}
\end{equation}
where $y_{min} = -2^{IL_{\bm{y}}}$, $y_{range} = 2^{IL_{\bm{y}}+1}$, and the modulo operator computes the remainder using floored division.

The quality of a solution in a given layer is evaluated in terms of Top-1 classification accuracy on a small validation dataset. If two quantization configurations within the same layer result in the same accuracy, we pick the solution that minimizes the Sum of Absolute Residuals (SAR). In other words, we compare the outputs of the floating-point layer to the outputs of the quantized layer, and pick the configuration that minimizes
\begin{equation}
	\sum_{i=1}^N |y_i - \hat{y_i}|
\end{equation} 
where $N$ equals the number of output samples within the current layer, and $y$ and $\hat{y}$ denote the floating-point and quantized output results, respectively.

\iffalse
 of the outputs results of the current layer between floating-point and fixed-point inference on the validation dataset:
\begin{equation}
	SAR(y,\hat{y}) = \sum_{i=1}^N |y_i - \hat{y_i}|
\end{equation}
\fi

\begin{figure}
\includegraphics[clip,width=90mm,keepaspectratio]{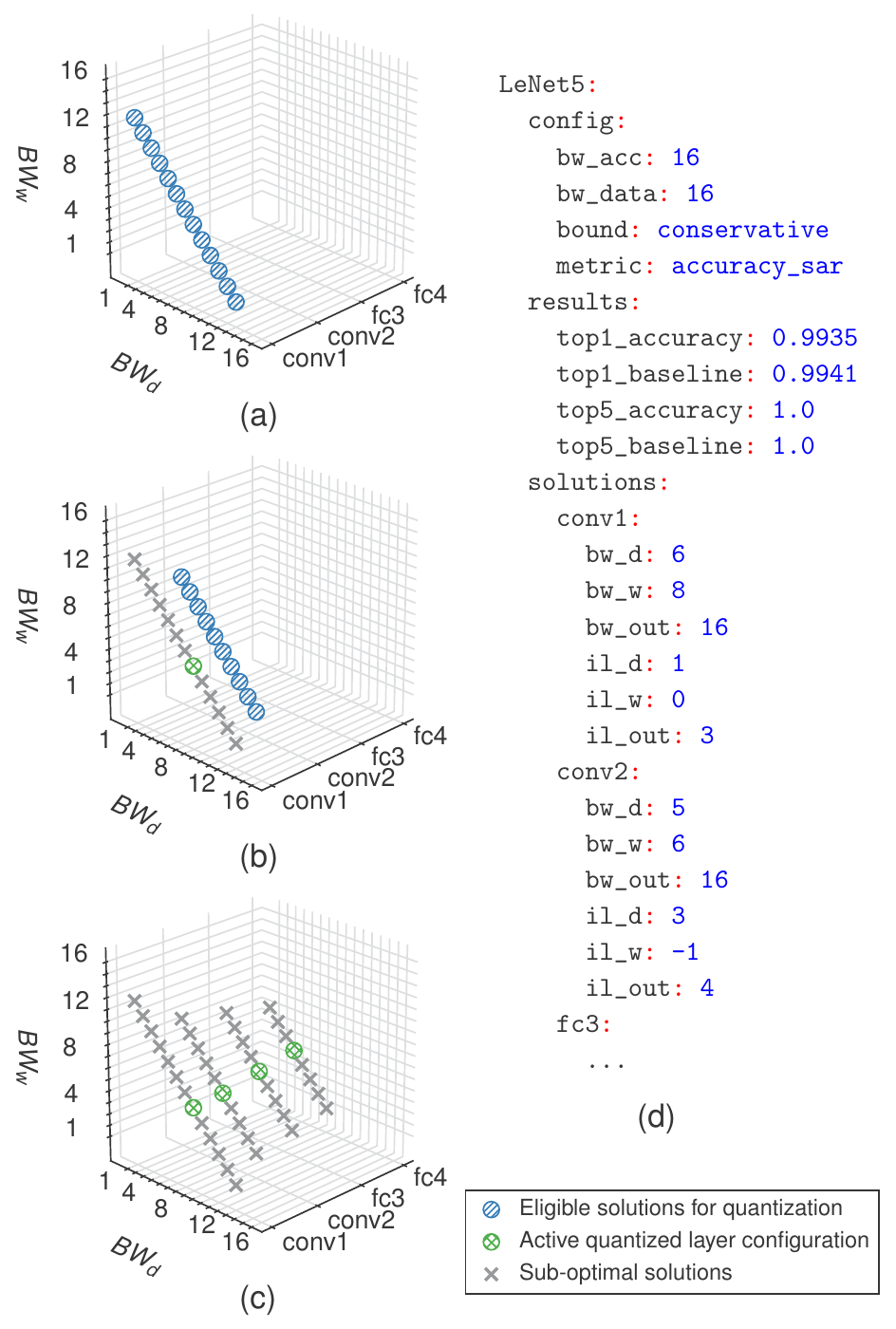}
\iffalse
\begin{minipage}[bt]{50mm}
\includegraphics[clip,width=50mm,keepaspectratio]{visualization/solution_space_plots/quantization_stages_all_transposed.pdf}
\end{minipage}%% 
\begin{minipage}[bt]{40mm}
%  \centering
\begin{adjustbox}{width=40mm}
\begin{lstlisting}[language=yaml,basicstyle=\linespread{1.1}\ttfamily,columns=fullflexible,keepspaces=true]
  LeNet5:
    config:
      bw_acc: 16
      bw_data: 16
      bound: conservative
      metric: accuracy_sar
    results:
      top1_accuracy: 0.9935
      top1_baseline: 0.9941
      top5_accuracy: 1.0
      top5_baseline: 1.0
    solutions:
      conv1:
        bw_d: 6
        bw_w: 8
        bw_out: 16
        il_d: 1
        il_w: 0
        il_out: 3
      conv2:
        bw_d: 5
        bw_w: 6
        bw_out: 16
        il_d: 3
        il_w: -1
        il_out: 4
      fc3:
        ...
\end{lstlisting}
\end{adjustbox}
  \end{minipage} 
\fi
\caption{Quantization procedure for a 4-layer network. (a) Analyse the chosen accumulator constraint for the first layer and test every eligible solution (blue dots). (b) Set the previous layer to best-performing solution (green dot) and continue quantization of the next layer. (c) Repeat the procedure until all layers are quantized. (d) Resulting output configuration.}
\label{Figure:quanti_procedure}
\end{figure}

The complete procedure for a 4-layer network is visualized in Figure \ref{Figure:quanti_procedure}. The quality of the final quantization result is measured in terms of Top-1 classification accuracy on a large separate test dataset. Note that different layers have a different number of available bits for $BW_{\bm{w}} + BW_{\bm{d}}$, depending on the data and parameter range and kernel size. In this example layer conv1 has 14 bits to distribute between $BW_{\bm{w}}$ and $BW_{\bm{d}}$, while layer conv2 has only 11 bits.
% dit misschien eerder
%We assume a fixed accumulator size ($BW_{acc}$) and maximum bit width for data and weights of every layer ($BW_{data}$). Having a different bit width for every fixed-point group is generally not feasible on general-purpose platforms.

\section{Fixed-point finetuning}
\label{sec:retraining}
This section presents a finetuning procedure to improve the quantization solutions from Section \ref{sec:heuristic}, without violating the accumulator constraints. The finetuning procedure is based on the idea of using a reduced-precision forward computation, in combination with a high-precision backwards pass\cite{Courbariaux2014, Gysel2018}. Dynamic fixed-point is used to update the scaling factors during training to prevent overflow\cite{Courbariaux2014, flexpoint}. We extend upon these works by including dynamic fixed-point scaling with accumulator constraints. After finetuning the fixed-point format is fixed, which allows for efficient model deployment.

The pre-trained floating-point model is initially set to the quantization solution that was found using the procedure of Section \ref{sec:heuristic}. During finetuning the SGD-optimizer computes the network loss over a mini-batch in the forward pass, calculates the loss contribution of all (quantized) weights in the backward pass, and finally updates the high-precision weights. The network loss is computed using the simulated fixed-point network. The derivatives of the discrete quantizers and wraparound operators (see Figure \ref{Figure:DP}) are estimated as an identity function. This gradient approximation function is also referred to as a Straight-Through Estimator\cite{DBLP:journals/corr/abs-1305-2982, DBLP:journals/corr/ZhouNZWWZ16}. The main intuition is that as long as the quantizer is fair, the average gradient at the quantization point will converge to the real gradient. Usually the identity function is clipped within the quantizer range to prevent the high-precision weights from exploding\cite{DBLP:journals/corr/ZhouNZWWZ16, hubara2017quantized}. However, we did not observe this behaviour in our benchmarks.

Updating the weights might result in accumulator overflow in any convolutional or fully-connected layer $l$ for any constraint except the pessimistic constraint, whose range is not data-dependent ($IL^l_{\bm{w}}$ and $IL^l_{\bm{d}}$ are fixed):
\begin{equation}
  IL^l_{acc} =
    \begin{cases}
      IL^l_{\bm{w}} + IL^l_{\bm{d}} + \lceil \log_2 K \rceil & \text{for pessimistic constraint}\\
      IL^l_{\bm{d}} + \lfloor \log_2 (R^l_{kernel}) \rfloor + 1 & \text{for conservative constraint}\\
      IL^l_{\bm{y}} & \text{for optimistic constraint}
    \end{cases}     
   \label{Eq:il_acc}  
\end{equation}

\noindent
During our initial experiments it became clear that accumulator overflow generally leads to catastrophic accuracy degradation. To resolve this, the accumulator range $IL^{l,batch}_{acc}$ of every quantized layer is computed during the forward computation of a mini-batch (see Figure \ref{Figure:DP}). If the accumulator overflows i.e.
\begin{equation}
IL^l_{acc} < IL^{l,batch}_{acc}
\label{Eq:il_ovf}  
\end{equation}
the bit width (and precision) of weights or input data will be reduced to increase the accumulator value range. This dynamic fixed-point scaling policy is similar to \cite{flexpoint} and slightly differs from \cite{Courbariaux2014}, where some overflow is tolerated. Deciding whether to reduce the bit width of weights or input data is not trivial, and might result in catastrophic accuracy degradation. Section \ref{sec:finetuning_heuristic} introduces a heuristic that makes this decision based on information that was collected during the quantization phase.

\subsection{Accumulator overflow resolution heuristic}
\label{sec:finetuning_heuristic}
During the quantization phase a number of solutions were tested to find the optimal quantization configuration. For each of these solutions the network loss on a small validation set was collected. We use this information during finetuning to decide whether to reduce the parameter or input data bit width. We base the decision on the assumption that
\begin{align*}
	&\textbf{if } Loss(BW^l_{\bm{d}}-1, BW^l_{\bm{w}}+1, l) \leq Loss(BW^l_{\bm{d}}+1, BW^l_{\bm{w}}-1, l) \\
	&\ \ \textbf{ then } Loss(BW^l_{\bm{d}}-1, BW^l_{\bm{w}}, l) \leq Loss(BW^l_{\bm{d}}, BW^l_{\bm{w}}-1, l)\\
	&\textbf{else } \text{wrap inequality sign}
\end{align*}
where the inequality of the first line is known from the quantization procedure, and the inequality of the second line is assumed. An example of this heuristic is visualized in Figure \ref{Figure:ft_procedure}. From the example follows that overflow is detected twice in the first epoch  within the same layer (for different mini-batches). This overflow is detected before wraparound (see Figure \ref{Figure:DP}), and can therefore be resolved without interrupting the finetuning procedure. A complete definition of the forward propagation during finetuning is depicted in Algorithm \ref{alg:1}.

\begin{figure}[htbp]
\centerline{\includegraphics[width=90mm,keepaspectratio]{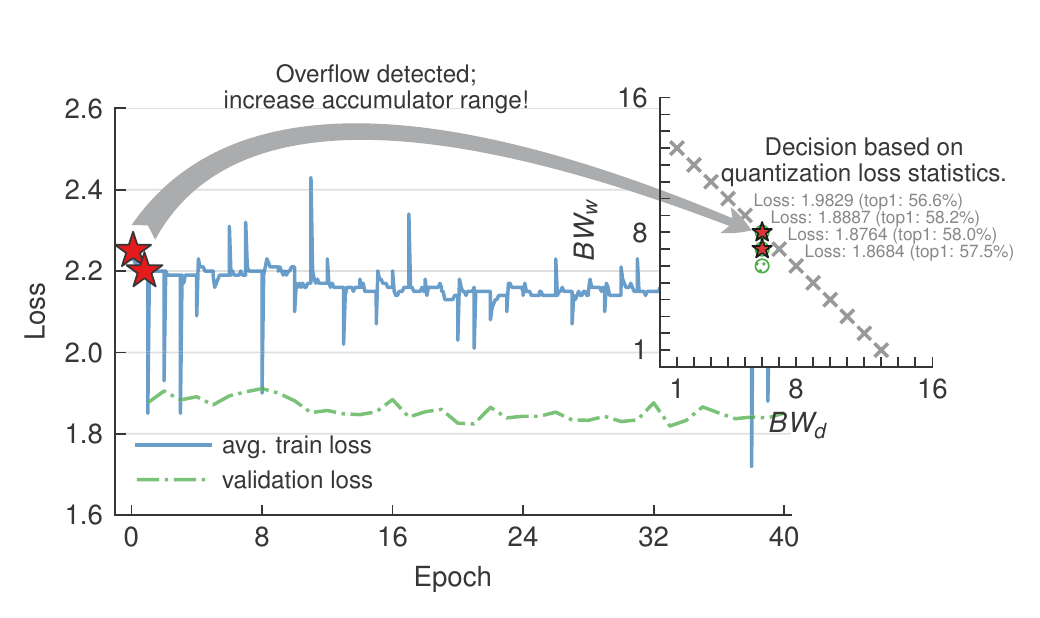}}
%%%\centerline{\includegraphics[width=90mm,keepaspectratio]{visualization/heuristic_v2-eps-converted-to.pdf}}
%  \begin{minipage}[b]{0.5\linewidth}
%    \centering
%    \includegraphics[width=0.95\linewidth]{visualization/figures_2/heuristic_1_of_2.eps}
%\vspace{-0.25cm}
%    \\(a)
%    %\vspace{4ex}
%  \end{minipage}%%
%  \begin{minipage}[b]{0.5\linewidth}
%    \centering
%    \includegraphics[width=0.95\linewidth]{visualization/figures_2/heuristic_2_of_2.eps}
%    \vspace{-0.2cm}
%    \\(b) 
%    %\vspace{4ex}
%  \end{minipage} 
\caption{Illustration of fixed-point finetuning procedure where the accumulator overflows 2 times in the same layer (red stars). When overflow is detected, the bit width of weights or input maps is being reduced based on the loss statistics that were collected during quantization.}
\label{Figure:ft_procedure}
\end{figure}

\begin{algorithm}
\caption{Forward computation of convolutional or fully-connected layer $l$ during fixed-point finetuning. $Loss$ returns the network loss that was found during layer-wise quantization.}\label{euclid}
\begin{algorithmic}[1]
\Procedure{Forward}{}
	\State{Compute layer output $\hat{y}$ knowing $\hat{d}$, $\hat{w}$\ and $\hat{b}$ using Equation \ref{eqn:quantized}}
	\State{Compute mini-batch accumulator range $IL^{l,batch}_{acc}$ using Equation \ref{Eq:il_acc}}
\While{$IL^l_{acc} < IL^{l,batch}_{acc}$}
	\If{$Loss(BW^l_{\bm{d}}-1, BW^l_{\bm{w}} + 1, l) \leq  Loss(BW^l_{\bm{d}}+1 , BW^l_{\bm{w}}-1, l)$}
		\State{Decrement $BW^l_{\bm{d}}$ and increment $IL^l_{acc}$}
	\Else
		\State{Decrement $BW^l_{\bm{w}}$ and increment $IL^l_{acc}$}
	\EndIf
\EndWhile
	\State{Compute $\hat{y}_{wrap}$ from $\hat{y}$ using Equation \ref{Eq:wrap}}
	\State{Quantize $\hat{y}_{wrap}$ using $BW^l_{out}$ and $IL^l_{out}$}
\EndProcedure
\end{algorithmic}
\label{alg:1}
\end{algorithm}

\iffalse
\begin{figure*}[bp]
\begin{minipage}{\textwidth}
    \centering % left, lower, right, upper
    \includegraphics[trim={0cm 9cm 0cm 2.75cm},clip,height=0.7\textheight]{visualization/solution_space_plots/quantization_stages_retraining.pdf}
\end{minipage}
\vspace{0.5cm}
\begin{minipage}{\textwidth}
    \centering % left, lower, right, upper
    \includegraphics[trim={0cm 4cm 0cm 0cm},clip,height=0.3\textheight]{visualization/solution_space_plots/finetuning_approach.png}
\end{minipage}
    \caption{Fine-tuning procedure for a 4-layer network. Top: Initialize network with quantization solution and start training. Middle: During training the accumulator of one or more layers might overflow. Using simulated fixed-point we can compute the magnitude of the overflow. Bottom: we reduce the accumulator precision by scaling weights or input data. We chose to reduce the precision of the group with the minimal input on loss. This information was already were collected during quantization.}
    \label{Figure:ft_procedure}
\end{figure*}
\fi

\iffalse
For simplicity we do only reduce the precision of weights, which implicitly reduces the precision of the accumulator, thereby preventing accumulator overflow within the same bit width:

where $BW_w$ and xxx are known and fixed (maybe add else case for constraint)
\fi

\section{Results and Evaluation} % Validation of quantization method
\iffalse
\documentclass[conference]{IEEEtran}
\IEEEoverridecommandlockouts
% The preceding line is only needed to identify funding in the first footnote. If that is unneeded, please comment it out.
\usepackage{cite}
\usepackage{amsmath,amssymb,amsfonts}
\usepackage{algorithmic}
\usepackage{graphicx}
\usepackage{textcomp}
\usepackage{multirow}
\usepackage{threeparttable}
\usepackage{adjustbox}
%\usepackage{dblfloatfix}    % To enable figures at the bottom of page
\usepackage{stfloats}
\usepackage{tabularx,booktabs}
\newcolumntype{C}{>{\centering\arraybackslash}X} % centered version of "X" type

\usepackage{soul}
\def\BibTeX{{\rm B\kern-.05em{\sc i\kern-.025em b}\kern-.08em
    T\kern-.1667em\lower.7ex\hbox{E}\kern-.125emX}}
    
\makeatletter
\def\endthebibliography{%
  \def\@noitemerr{\@latex@warning{Empty `thebibliography' environment}}%
  \endlist
}
\makeatother
\begin{document}
\fi

In this section we will evaluate the proposed quantization and finetuning method on 3 popular CNN benchmarks for image classification; LeNet5 (MNIST), All-CNN-C (CIFAR10) and AlexNet (ILSVRC2012). In particular, we evaluate the effectiveness of the different accumulator constraints in comparison to floating-point baselines, including finetuning. We continue by investigating the optimality of the accumulator constraints in comparison to an optimistic lower bound. We conclude by an evaluation on a real integer-based platform.

All quantization experiments were performed on an Ubuntu 17.10 machine with 16GB RAM, an Intel-i5 7300HQ, and a GTX1050 with 4GB VRAM. The quantization procedure, as presented in Section \ref{sec:heuristic} and \ref{sec:retraining} was implemented in PyTorch 0.3.1 and accelerated by the GPU using CUDA 8.0 with cuDNN v6. Quantized layers were simulated by reducing the precision and range of data and weights before and after the kernel computation. Overflow behaviour in the accumulator was simulated as well. For simplicity we do only quantize convolutional and fully-connected layers. Activation, subsampling, and normalization layers are not quantized. However, these layers do generally not have a significant impact on the total CNN workload.

\subsection{Baseline Networks}
\subsubsection{LeNet5 (MNIST)} We evaluate our quantization procedure first on a small 5-layer LeNet5(-like) network. This 5-layer CNN is trained on the 10-class MNIST dataset for handwritten digit classification. The network structure is depicted in Table \ref{Table:LeNet5_1}. All layers (except the last) are followed up by a ReLU activation function. Convolutional layers are also sub-sampled by a 2$\times$2 Max-pooling filter with stride 2. 200 images were sampled from the official MNIST training set for quantization. The final solution is verified on the official validation set. For fixed-point finetuning we use a mini-batch SGD optimizer with learning rate equal to 1e-4, momentum of 0.9 and a L2 weight regularization term of 5e-4. We finetune for 20 epochs and validate the resulting models on the official validation set.
\begin{table}[h]
\centering
\caption{LeNet5 network structure.}
\begin{tabular}{lll}
\toprule
Layer & Kernel size (+ bias) & Output size \\
\midrule
input & -- & $28\times 28\times 1$ \\
conv1 & $5\times 5 + 1$ & $24\times 24\times 16$ \\
conv2 & $5\times 5\times 16 + 1$ & $8\times 8\times 32$ \\
fc3   & $4\times 4\times 32 + 1$ & $512$ \\
fc4   & $512 + 1$ & $10$ \\
\bottomrule
\label{Table:LeNet5_1}
\end{tabular}
\vspace{-0.5cm}
\end{table}

\subsubsection{All-CNN-C (CIFAR10)} This 9-layer network\cite{DBLP:journals/corr/SpringenbergDBR14} is trained on the popular CIFAR-10 dataset. We use the pre-trained model from the Nervana model zoo\footnote{https://gist.github.com/nervanazoo}. Similar to LeNet5, 200 images were randomly sampled from the official training dataset for quantization. The final solution was tested on the official test dataset. This network contains layers with larger kernels (e.g. \textgreater 1000 multiply-accumulations) and is therefore potentially harder to quantize for a platform with narrow accumulators. For fixed-point retraining all training settings were kept identical to the floating-point reference, except for the learning rate, which was set at 5e-3 and decreased to 5e-6 by dividing by 10 every 5 epochs. We finetune for 20 epochs and validate the final solution on the official validation set.

\subsubsection{AlexNet (ILSVRC2012)} To investigate the limits of our quantization approach, we also quantized the well-known 8-layer AlexNet CNN\cite{Krizhevsky2014}. This network is trained on the difficult 1000-class ImageNet ILSVRC2012 dataset. We use the pre-trained model from the PyTorch model zoo\footnote{https://github.com/pytorch/vision}. 1000 images were randomly sampled from the training set for optimization. The quantized solution was tested on the official 50K validation set. Similar to All-CNN-C, this network contains several layers with large kernels. For fixed-point finetuning all training settings were kept identical to the floating-point reference, except for the learning rate, which was set at 5e-4 and decreased to 5e-7 by dividing by 10 every 10 epochs. We retrain for 40 epochs on 10\% of the training set and validate the final solution on the official validation set.

\subsection{Maximizing accuracy for different accumulator sizes}
\begin{figure*}[h]
    \centering
    \centerline{\includegraphics[width=170mm,keepaspectratio]{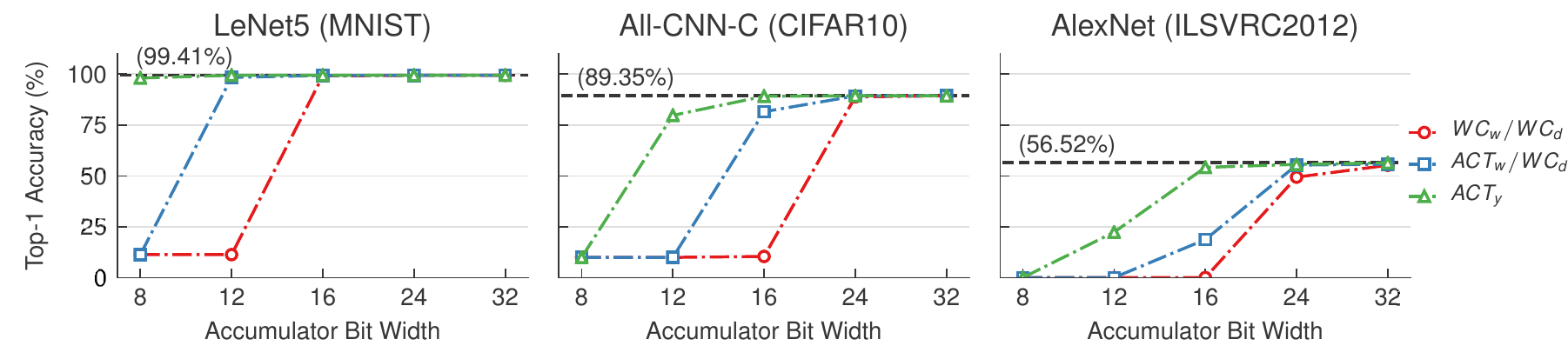}}
    \caption{Quantization results for various accumulator bit widths for the 3 constraints. The dashed line indicates the floating-point reference accuracy. It is clearly visible that the most relaxed constraint ($ACT_y$) achieves the highest accuracy.}
    \label{Figure:bw_acc1}
\end{figure*}

Figure \ref{Figure:bw_acc1} visualizes the quantization results for different accumulator sizes. For this experiment the maximum data bit width $BW_{data}$ is set equal to $BW_{acc}$. For all three benchmarks a 32-bit accumulator yields no significant loss in Top-1 classification accuracy, compared to the floating-point baseline. 

\noindent
As expected, the $WC_w/WC_d$ constraint performs worse than the other constraints, since its pessimistic accumulator range estimate limits the precision within the kernel computation. Although the $ACT_y$ constraint is potentially unsafe, it matches or outperforms the other two constraints on all benchmarks for different accumulator bit widths. For the small LeNet5 benchmark the accumulator bit width can be reduced to 12-bit, before the relative error increases by more than 20\%. For the other benchmarks the accumulator bit width can be reduced to 16-bit within a 5\% relative error penalty.

\subsection{Maximizing accuracy for various data bit widths}
\label{sec:exp3}
In the previous experiment the maximum data bit width $BW_{data}$ was fixed to $BW_{acc}$. However, to satisfy the accumulator constraints, $BW_{\bm{w}}$ and $BW_{\bm{d}}$ are generally set to values way smaller than $BW_{data}$. As a consequence, many bits are left unused on a typical general-purpose platform that only supports several integer data types. Therefore we have investigated the obtainable classification accuracy for different choices for $BW_{data}$. In this experiment we only used the most optimistic $ACT_y$ constraint. The results are listed in Table \ref{Table:3}. 
\begin{table*}
\caption{Quantization results (Top-1 accuracy) for restricted data and accumulator bit width before fixed-point retraining (using constraint $ACT_{y}$). Percentages between brackets indicate the absolute accuracy improvement after finetuning (improvements $\leq$0.5\% are not shown).}
%\begin{adjustbox}{width=1\textwidth}
\centering
\begin{tabular}{llccc}
\toprule
$BW_{acc}$ & Max. $BW_{data}$ &  LeNet5 &  All-CNN-C & \multicolumn{1}{c}{AlexNet} \\
\midrule
Float32 & Float32   &   99.4\%          & 89.4\%          & 56.5\%  \\
32 & 32 &          99.4\% &          89.4\% &           56.0\% \\
   & 24 &          99.4\% &          89.4\% &           56.5\% \\
   & 16 &          99.4\% &          89.4\% &           56.5\% \\
   & 12 &          99.4\% &          89.4\% &           56.5\% \\
   & 8  &          99.4\% &  88.8\% (+0.7\%) &   53.9\% (+1.2\%) \\
24 & 24 &          99.4\% &          89.4\% &           56.4\% \\
   & 16 &          99.4\% &          89.4\% &           56.4\% \\
   & 12 &          99.4\% &          89.4\% &           56.4\% \\
   & 8  &          99.4\% &          88.8\% &   54.0\% (+0.9\%) \\
16 & 16 &          99.4\% &          89.1\% &   54.6\% (+0.6\%) \\
   & 12 &          99.4\% &          89.1\% &   53.8\% (+1.2\%) \\
   & 8  &          99.4\% &          89.0\% &   53.6\% (+0.9\%) \\
12 & 12 &          99.4\% &  79.8\% (+5.2\%) &  21.4\% (+22.2\%) \\
   & 8  &          99.4\% &  81.1\% (+3.3\%) &  21.5\% (+17.1\%) \\
   & 4  &          98.6\% &  11.5\% (-1.4\%) &            0.2\% \\
8  & 8  &  98.1\% (+0.6\%) &  -- &            -- \\
   & 4  &  92.5\% (+5.6\%) &  -- &           -- \\
\bottomrule
\end{tabular}
%\end{adjustbox}
\label{Table:3}
\end{table*}

For a sufficiently large accumulator (i.e. 32-bit), 16-bit data yields no loss in accuracy for any of the benchmarks. This is in line with other works\cite{DBLP:journals/corr/GyselMG16, Guo2017}. From the results follows that the accumulator of all benchmarks can be reduced to 24-bit and 12-bit data without a performance penalty. For 8-bit data types, the classification accuracy of AlexNet is expected to drop by 1--2\%\cite{Gysel2018}. This penalty only increases marginally when the accumulator is reduced from 32-bit to 16-bit. Reducing to accumulator to 8-bit only reduces the bit width of weights and input data to only several bits. For the All-CNN-C (Cifar10) and AlexNet (ILSVRC2012) benchmarks the resulting quantization solutions perform as well as random guessing, and the finetuning procedure is not able to recover from this large accuracy degradation. This is expected, since competing solutions that succeed on $<$8-bit quantization typically take special measures, such as only quantizing inputs or weights\cite{journals/corr/RastegariORF16, DBLP:journals/corr/LaiSC17}, exploiting a non-linear or assymetric fixed-point number format\cite{tensorflow2015-whitepaper}, or keep the first and last layer in high-precision\cite{journals/corr/RastegariORF16, DBLP:journals/corr/ZhouNZWWZ16}. 
%Fine-grained channel-wise quantization and assymetric quantizers might also improve the results.

\subsection{Further accuracy improvements with fixed-point finetuning}
To reduce the accuracy penalty even further, we applied finetuning to a selected number of quantization solutions. We start with the quantization solutions from Section \ref{sec:exp3} and apply the retraining procedure as explained in Section \ref{sec:retraining} to the floating-point reference model. 

We first demonstrate the effectiveness of our accumulator overflow resolution heuristic by finetuning a selection of quantization solutions with narrow accumulators. For comparison, a variety of overflow resolution heuristics were tested as well: Never reduce $BW_{\bm{w}}$/$BW_{\bm{d}}$, always reduce $BW_{\bm{d}}$, always reduce $BW_{\bm{w}}$, and our proposed heuristic (Algorithm \ref{alg:1}). It follows from the results in Table \ref{Table:7} that the simple heuristics do not work in all cases, and that the proposed heuristic works well in preventing catastrophic accuracy degradation during finetuning. One interesting observation is that always reducing parameter bit width fails  only in one extreme case (i.e. LeNet5 with 4-bit data and 8-bit accumulator), but generally regains more accuracy than the proposed heuristic.
\begin{table*} %[!b]
\caption{Accumulator overflow resolution heuristic compared to several simpler alternatives.}
%\begin{adjustbox}{width=1\textwidth}
\centering
\begin{tabular}{lllllll}
\toprule
{} & \multicolumn{2}{l}{LeNet5} & \multicolumn{2}{l}{All-CNN-C} & \multicolumn{2}{l}{AlexNet} \\
($BW_{acc}$, Max. $BW_{data}$)  & (8, 4) & (8, 8) &   (12, 8) & (12, 12) & (16, 8) & (16, 16) \\
\midrule
Baseline (Float32)             &  99.4\% &  99.4\% &     89.4\% &    89.4\% &   56.5\% &    56.5\% \\
Before finetuning                  &  92.5\% &  98.1\% &     81.1\% &    79.8\% &   53.6\% &    54.6\% \\
\hline
Never reduce $BW_{\bm{w}}$/$BW_{\bm{d}}$  &  {\color{red}19.6\%} &  98.4\% &   \textbf{86.4}\% &    \textbf{86.9\%} &    {\color{red}0.1\%} &     {\color{red}0.1\%} \\
Always reduce $BW_{\bm{d}}$    &  \textbf{98.3\%} &  {\color{red}11.4\%} &     {\color{red}71.8\%} &    84.6\% &   54.2\% &    54.2\% \\
Always reduce $BW_{\bm{w}}$    &  {\color{red}11.4\%} &  \textbf{98.7\%} &     \textbf{86.4}\% &    86.5\% &  \textbf{54.8\%} &    55.1\% \\
Proposed (Algorithm \ref{alg:1})             &  \textbf{98.3\%} &  98.6\% &     84.2\% &    85.0\% &   54.5\% &    \textbf{55.2\%} \\
\bottomrule
\end{tabular}
%\end{adjustbox}
\label{Table:7}
\end{table*}

Additional finetuning results with the proposed accumulator overflow heuristic are shown in Table \ref{Table:3}. LeNet5 with a 4-bit data bus and an 8-bit accumulator is able to regain most of its lost precision. All-CNN-C with a 12-bit accumulator is able to regain over 5\% of its lost accuracy. For AlexNet the finetuning procedure is able to regain a respectable amount of accuracy in especially the 8-bit and 12-bit designs with small accumulators ($\leq$16-bit).

Overall these experiments demonstrate that large CNNs can run effectively on platforms without support for wide accumulators. Additionally, it has been shown that the proposed quantization heuristic for layer-wise optimization in combination with fixed-point finetuning obtains good solutions for narrow accumulators for a variety of maximum data bit widths.

\subsection{Analysis of accumulator constraints}
For all previous experiments the accumulator range was chosen such that the chances of the accumulator overflowing are minimized (or completely avoided). However, some overflow might be tolerable. This experiment aims to provide insights on an optimistic lower bound on the minimum accumulator range. 

\begin{figure}[htbp]
\centerline{\includegraphics[width=90mm,keepaspectratio]{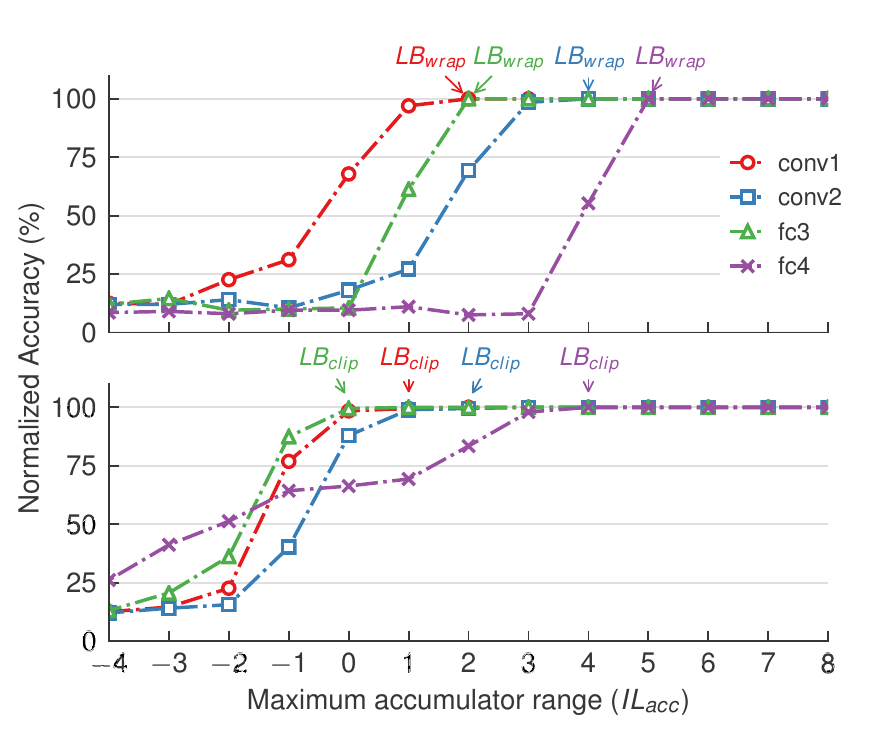}}
\vspace{-0.3cm}
\caption{LeNet5 normalized Top-1 accuracy on small (200-image) dataset when the accumulator range (i.e. $IL_{acc}$) of a single layer is restricted. Overflow behaviour is either wrap-around (top) or clipping (bottom). For every layer the minimum range before accuracy collapses is indicated.}
\label{Figure:LeNet5_3}
\vspace{-0.2cm}
\end{figure}

%To determine the quality of every constraint, we continue to investigate the minimum required accumulator range per layer. 
We start with a high-precision network and reduce the representable accumulator range (i.e. $IL_{acc}$) of a single layer. We then forward a small dataset (200 images) through the modified network and observe the Top-1 classification accuracy. Two types of overflow-behaviour were considered: wrap-around and clipping. The results for LeNet5 are depicted in Figure \ref{Figure:LeNet5_3}. It follows from the figure that the accuracy drops very steeply when the accumulator has insufficient range. Clipping seems to delay this accuracy breakdown point by at least 1--2 bits of additional accumulator range.

\noindent
To compare the minimal required accumulator range to the solutions that were found by the accumulator constraints from Section \ref{sec:qwna}, two metrics have been defined: $LB_{wrap}$ and $LB_{clip}$. $LB_{wrap}$ denotes the minimum integer length of the accumulator for which the normalized classification accuracy is still above 99\% for the wrap-around case. Similarly, $LB_{clip}$ denotes the minimum integer length for the case where clipping is used. 

For the LeNet5 benchmark the required integer lengths are listed in Table \ref{Table:LeNet5_2}. The optimistic $ACT_y$ constraint is generally very close to the point where the network accuracy starts to collapse ($LB_{wrap}$). If the accumulator would clip instead of wrap-around, we could potentially reduce $IL_{acc}$ by another 1 or 2 bits. The two safe constraints are too pessimistic, which results in suboptimal network accuracy for accumulators below 32-bit (see Figure \ref{Figure:bw_acc1}).

\begin{table}[h]
\centering
\caption{Required accumulator range ($IL_{acc}$) for LeNet5 benchmark with different constraints.}
\begin{adjustbox}{max width=\textwidth}
\begin{threeparttable}
\begin{tabular}{llllll}
\toprule
Layer & $WC_w/WC_d$ & $ACT_w/WC_d$\tnote{a} & $ACT_{y}$ & $LB_{wrap}$\tnote{b} & $LB_{clip}$\tnote{b} \\
\midrule
conv1 & 6  & 4 & 3 & 2 & 1 \\
conv2 & 11 & 8 & 4 & 4 & 2 \\
fc3   & 12 & 8 & 4 & 2 & 0 \\
fc4   & 12 & 8 & 5 & 5 & 4 \\
\bottomrule
\end{tabular}
\begin{tablenotes}
\item [a] \scriptsize Quantization errors due to rounding were ignored for this experiment.
\item [b] \scriptsize Smallest integer length where the normalized accuracy is still above 99\% (see Figure \ref{Figure:LeNet5_3}).
\end{tablenotes}
\end{threeparttable}
\end{adjustbox}
\label{Table:LeNet5_2}
\end{table}

\noindent
% Quantization errors were ignored for constraint 2
The same analysis was applied to the other benchmarks for which results are summarized in Table \ref{Table:OtherNets}. Depending on the layer, the required accumulator range for the $ACT_y$ constraint could potentially be reduced by 1--2 bits. However, finding these quantization solutions is non-trivial and boils down to testing more options. Finding these solutions could potentially lead to better results for even smaller accumulators bit widths.
\begin{table}[h]%[tp]
\caption{Required accumulator range ($IL_{acc}$) of other benchmarks with different constraints.}
\label{Table:OtherNets}
\centering
%    \begin{minipage}{.55\linewidth}
%%\begin{adjustbox}{width=0.4\textwidth}
\begin{tabular}{llllll}
\toprule
Layer & $WC_w/WC_d$ & $ACT_w/WC_d$\tnote{a} & $ACT_{y}$ & $LB_{wrap}$\tnote{b} & $LB_{clip}$\tnote{b} \\
\midrule
conv1 &        8 &        6 &        4 &       1 &       0 \\
conv2 &       13 &       10 &        6 &       3 &       2 \\
conv3 &       14 &       11 &        8 &       5 &       4 \\
conv4 &       16 &       13 &        9 &       7 &       5 \\
conv5 &       18 &       15 &        9 &       8 &       6 \\
conv6 &       18 &       14 &        8 &       6 &       5 \\
conv7 &       16 &       14 &        9 &       7 &       6 \\
conv8 &       16 &       13 &        9 &       7 &       6 \\
conv9 &       18 &       14 &       11 &      10 &       9 \\
\bottomrule
\end{tabular}\vspace{0.25cm}
%%\end{adjustbox}
\\(a) All-CNN-C
\ \\
\ \\
%    \end{minipage}%
%    \begin{minipage}{.55\linewidth}
%      \centering
%%\begin{adjustbox}{width=0.4\textwidth}
\begin{tabular}{llllll}
\toprule
Layer & $WC_w/WC_d$ & $ACT_w/WC_d$\tnote{a} & $ACT_{y}$ & $LB_{wrap}$\tnote{b} & $LB_{clip}$\tnote{b} \\
\midrule
conv1 &       11 &        8 &        6 &       5 &       5 \\
conv2 &       19 &       13 &        8 &       8 &       5 \\
conv3 &       18 &       14 &        8 &       7 &       5 \\
conv4 &       18 &       14 &        7 &       6 &       4 \\
conv5 &       17 &       13 &        7 &       6 &       4 \\
fc6   &       17 &       13 &        7 &       6 &       4 \\
fc7   &       16 &       13 &        7 &       6 &       4 \\
fc8   &       17 &       13 &        6 &       6 &       5 \\
\bottomrule
\end{tabular}
%%\end{adjustbox}
\vspace{0.25cm}
\\(b) AlexNet
%\end{adjustbox}
%    \end{minipage}
\end{table}

\newpage
\subsection{Evaluation on an ARM processor}
\label{sec:arm_proc}
For validation purposes and demonstration of the effectiveness of accumulator-constrained quantization, we have ported the obtained quantization solutions to an ARM Cortex A53 (Raspberry Pi 3) processor with a NEON coprocessor without wide accumulator support. This coprocessor operates on 128-bit vector registers and provides a variable degree of SIMD-parallelism, based on the input data types. It supports 16$\times$8-bit, 8$\times$16-bit or 4$\times$32-bit integer operations in parallel. Floating-point support is limited to 4$\times$32-bit parallel operations. These types of vector processors are very common for imaging and computer vision applications.

The floating-point Pytorch models and preprocessed benchmark images were converted to fixed-point, and stored in the closest integer data type. All relevant NN layers were implemented in C++. It follows from the results in Table \ref{Table:10} that the simulated fixed-point Pytorch results approximately match the integer-based fixed-point solutions. All-CNN-C and AlexNet do not function with 8-bit accumulators, and are therefore not shown.
\begin{table*}[h]
\caption{Results of finetuned networks (Pytorch vs C++) on official validation sets. For AlexNet a random 500-image sample was used.}
%\begin{adjustbox}{width=1\textwidth}
\centering
\begin{tabular}{llcccccc}
\toprule
   &    &  \multicolumn{2}{c}{LeNet5} &  \multicolumn{2}{c}{All-CNN-C} & \multicolumn{2}{c}{AlexNet} \\
$BW_{acc}$ & Max. $BW_{data}$  & Pytorch      & C++    & Pytorch      & C++   & Pytorch & C++   \\
\midrule
Float32 & Float32 &        99.4\% &        99.4\% &          89.4\% &        89.4\% &              54.6\% &        54.6\% \\
32 & 32 &        99.4\% &        99.4\% &          89.5\% &        89.5\% &              54.4\% &        54.6\% \\
  & 16 &        99.4\% &        99.4\% &          89.8\% &        89.8\% &              53.8\% &        53.8\% \\
  & 8 &        99.4\% &        99.4\% &          89.3\% &        89.4\% &              53.0\% &        53.0\% \\
16 & 16 &        99.4\% &        99.4\% &          89.5\% &        89.5\% &              53.0\% &        53.0\% \\
  & 8 &        99.4\% &        99.4\% &          89.2\% &        89.1\% &              51.8\% &        52.4\% \\
8 & 8 &        98.6\% &        98.7\% &          -- &        -- &               -- &         -- \\
  & 4 &        98.2\% &        98.5\% &          -- &         -- &               -- &         -- \\
\bottomrule
\end{tabular}
%\end{adjustbox} % should match
\label{Table:10}
\end{table*}

\noindent
%All Halide (C++) mappings make extensive use of multithreading and vectorization. Schedules between mappings are very similar across all accumulator sizes. The only change is the vector width, which changes from 4 (float and int32) to 8 (int16) or even 16 (int8).

\begin{figure}[htbp]
\centering
\includegraphics[width=90mm,keepaspectratio]{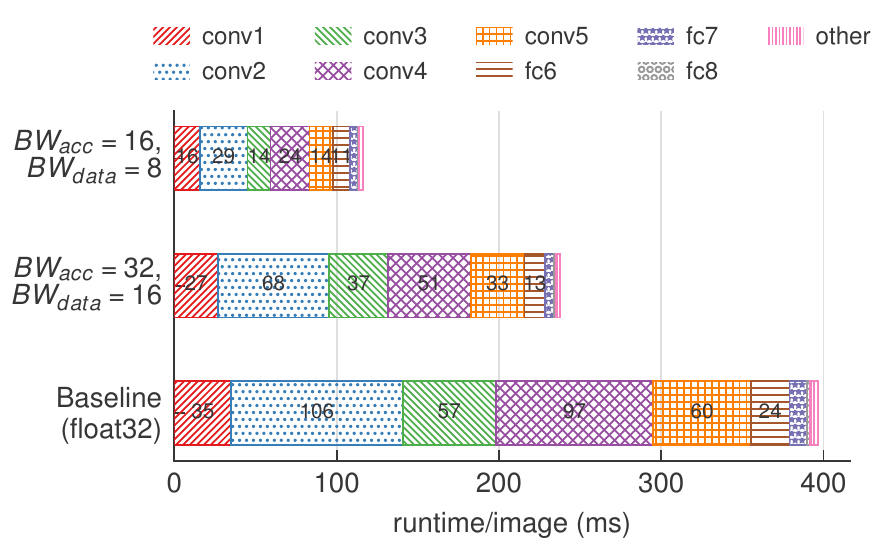}
%\includegraphics[width=90mm,keepaspectratio]{visualization/alexnet_runtime-eps-converted-to.pdf}
%\vspace{-0.3cm}
\caption{Performance breakdown of individual layers of AlexNet (over 10 runs) on ARM with varying degree of SIMD-parallelism (depending on accumulator size $BW_{acc}$).}
\label{Figure:alexnet_runtime}
%\vspace{-0.2cm}
\end{figure}

We continue by evaluating the obtainable speedup by exploiting smaller accumulators in combination with higher degrees of SIMD-parallelism. All benchmarks are manually optimized for different accumulator sizes and make extensive use of multithreading and vectorization (i.e. all Float32 benchmarks reach approximately 30\% of peak GFLOPs compared to the MP-MFLOPS NEON benchmark\cite{Longbottom2016}). We find that over 90\% of the runtime in the floating-point baseline is spend in convolution and fully-connected layers for AlexNet, as is shown in Figure \ref{Figure:alexnet_runtime}. 
Reducing the accumulator (and multipliers) from 32-bit to 16-bit leads to throughput improvements with almost ideal scaling by exploiting wider vector-instructions. Except for the vector-width, the schedule remains unchanged. The runtime discrepancy between the baseline and 32-bit integer mapping may be attributed to the reduced floating-point throughput of the NEON coprocessor. 

\begin{figure}[htbp]
\centering
\centerline{\includegraphics[width=170mm,keepaspectratio]{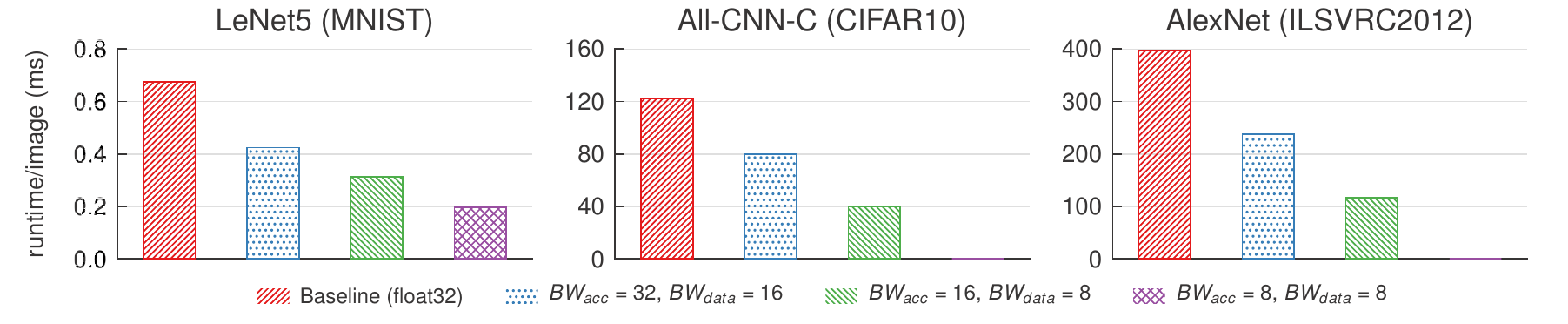}}
%\centerline{\includegraphics[width=170mm,keepaspectratio]{visualization/network_performance-eps-converted-to.pdf}}
%\vspace{-0.3cm}
\caption{Average throughput (over 10 runs) on ARM with varying degree of SIMD-parallelism (depending on accumulator size $BW_{acc}$).}
\label{Figure:network_performance}
%\vspace{-0.2cm}
\end{figure}

\noindent
Similar throughput improvements can be achieved by the other benchmarks, as depicted in Figure \ref{Figure:network_performance}. Performance scaling on the LeNet5 benchmark is slightly less due to SIMD underutilization caused by small feature maps.

\section{Conclusions}
This paper presents a new quantization method for integer-based platforms without support for wide kernel accumulators. Two constraints to maximize the bit width of weights and input data for a given accumulator size are introduced. Using these constraints a layer-wise quantization heuristic for finding good fixed-point approximations is proposed. Only solutions that fully utilize the available accumulator bits are tested. We have evaluated our quantization method on three popular CNNs for image classification, and have demonstrated that 16-bit accumulators are sufficient for large CNNs. The results show that the narrow accumulators with our quantization technique in combination with finetuning can still deliver good classification performance. The mapping on an ARM processor with vector extensions reveals that near-ideal throughput scaling is possible by using 16-bit accumulators. In future research we could exploit these findings for designing efficient accelerators and processors for NNs and extend our approach to other application domains.

%\section*{References}
%\bibliographystyle{IEEEtran}
\bibliographystyle{model1-num-names}
\bibliography{quantization_paper}
%\nocite{*}

\end{document}